\documentclass[twoside]{article}

\usepackage[accepted]{aistats2026} 

\usepackage[round]{natbib}
\usepackage{amsmath,amssymb}
\usepackage{booktabs}
\usepackage{xcolor}
\usepackage{multirow}
\usepackage{multicol}
\usepackage{graphicx}
\usepackage{algorithm}
\usepackage{algpseudocode}

\newcommand{\bg}[1]{\boldsymbol{#1}} 
\newcommand{\bm}[1]{\mathbf{#1}} 

\newcommand\T{{\mathpalette\raiseT\intercal}}
\newcommand\raiseT[2]{%
\setbox0\hbox{$#1{#2}$}\raise\dp0\box0}

\usepackage[hidelinks]{hyperref}

\usepackage{pifont} 
\newcommand{\cmark}{\text{\ding{51}}}
\newcommand{\xmark}{\textcolor{lightgray}{\ding{55}}}

\begin{document}

\runningtitle{Time Series Forecasting with Hahn Kolmogorov-Arnold Networks}
\runningauthor{Md Zahidul Hasan, A. Ben Hamza,  Nizar Bouguila}

\twocolumn[
\aistatstitle{Time Series Forecasting with Hahn Kolmogorov-Arnold Networks}
\aistatsauthor{Md Zahidul Hasan \And A. Ben Hamza \And  Nizar Bouguila}
\aistatsaddress{Concordia Institute for Information Systems Engineering\\Concordia University, Montreal, QC, Canada}
]

\begin{abstract}
Recent Transformer- and MLP-based models have demonstrated strong performance in long-term time series forecasting, yet Transformers remain limited by their quadratic complexity and permutation-equivariant attention, while MLPs exhibit spectral bias. We propose HaKAN, a versatile model based on Kolmogorov-Arnold Networks (KANs), leveraging Hahn polynomial-based learnable activation functions and providing a lightweight and interpretable alternative for multivariate time series forecasting. Our model integrates channel independence, patching, a stack of Hahn-KAN blocks with residual connections, and a bottleneck structure comprised of two fully connected layers. The Hahn-KAN block consists of inter- and intra-patch KAN layers to effectively capture both global and local temporal patterns. Extensive experiments on various forecasting benchmarks demonstrate that our model consistently outperforms recent state-of-the-art methods, with ablation studies validating the effectiveness of its core components. 
\end{abstract}

\section{INTRODUCTION}
Time series forecasting is widely used as a critical tool in diverse domains ranging from retail, energy and transportation to healthcare and finance~\citep{Wang2025Traffic,Zhang2025MultiPeriod}. However, this task poses significant challenges due to the need to effectively capture complex temporal patterns and long-range dependencies, while maintaining computational efficiency.

Recent advances in multivariate time series forecasting have explored Transformer- and MLP-based models to address these challenges. Transformer-based methods~\citep{Zhou2021Informer,Wu2021Autoformer,Zhou2022FEDformer,Liu2024iTransformer} rely on attention mechanisms to capture long-range dependencies, with simple strategies such as channel independence and patching~\citep{Nie2023PatchTST} contributing to improved efficiency and predictive performance. However, Transformers often suffer from high computational complexity, quadratic in sequence length, and their permutation-equivariant attention also contradicts the causal nature of time series data. On the other hand, MLP-based methods~\citep{Zeng2023DLinear,Das2023TiDe} offer a computationally lighter alternative by using linear layers to model temporal patterns, often incorporating the channel independence strategy to capture channel-specific patterns. Despite their efficiency, MLPs exhibit spectral bias~\citep{Rahaman2019Bias}, which limits their ability to model high-frequency components in time series, and struggle with capturing non-linear temporal dynamics due to their reliance on linear transformations, leading to suboptimal performance on datasets, where non-linear patterns dominate. More recently, Kolmogorov-Arnold Networks (KANs)~\citep{Liu2024KAN, Wang2024KANs} have emerged as a viable alternative to MLPs, offering a promising solution to the aforementioned limitations by replacing fixed activation functions with learnable functions, parameterized using splines. Rooted in the Kolmogorov-Arnold representation theorem~\citep{Braun2009KANTH, Johannes2021KANTH}, KANs are interpretable and mitigate spectral bias by enabling flexible function approximation, allowing the model to capture both low- and high-frequency components in the data~\citep{Wang2024KANs}. This adaptability makes KANs particularly well-suited for long-term forecasting, where diverse temporal patterns, ranging from short-term fluctuations to long-term trends, must be modeled accurately and efficiently.

\textbf{Proposed Work and Contributions.}\quad We propose Hahn Kolmogorov-Arnold Network (HaKAN)\footnote{Code: \url{https://github.com/zadidhasan/HaKAN}}, a novel framework for multivariate long-term time series forecasting, where each KAN layer is parameterized using Hahn Polynomials~\citep{Koekoek2010Poly}, enabling flexible and efficient function approximation. Unlike Transformer-based models, HaKAN avoids the computational overhead of attention mechanisms by using inter- and intra-patch KAN layers to model temporal relationships. Compared to MLP-based models, HaKAN employs learnable activation functions based on Hahn polynomials to capture non-linear temporal dynamics, overcoming the limitations of linear transformations. HaKAN also incorporates channel independence, patching, and a bottleneck structure to enhance robustness and efficiency, making it well-suited for diverse forecasting datasets and across various prediction horizons. The proposed framework combines the flexibility of KANs with a hierarchical patch-based design, enabling our model to capture both global and local temporal patterns while maintaining interpretability through learnable activation functions. The key contributions of this paper can be summarized as follows: (i) We introduce HaKAN, an effective framework for multivariate long-term time series forecasting that leverages the expressive power of KANs; (ii) we design a novel architecture featuring an Hahn-KAN block that integrates inter- and intra-patch KAN layers to effectively capture both global and local temporal patterns, respectively; and (iii) we demonstrate through extensive experiments that our model consistently outperforms strong baselines.

\section{RELATED WORK}\label{sec:related_work}
\textbf{Transformer-based Models.}\quad A sizable body of research has focused on designing Transformer-based methods for long-term time series forecasting~\citep{Liu2021Pyraformer,Zhou2021Informer,Wu2021Autoformer,Zhou2022FEDformer,Liu2024iTransformer,Nie2023PatchTST}. For instance, Informer~\citep{Zhou2021Informer} enhances Transformer efficiency with a ProbSparse self-attention mechanism, self-attention distilling, and a generative decoder. Autoformer~\citep{Wu2021Autoformer} introduces a decomposition architecture with an auto-correlation mechanism that leverages series periodicity for dependency discovery and representation aggregation. FEDformer~\citep{Zhou2022FEDformer} integrates seasonal-trend decomposition with Fourier and Wavelet transforms to capture global time series characteristics, while iTransformer~\citep{Liu2024iTransformer} inverts the traditional Transformer architecture by embedding entire time series of individual variates as tokens, using attention to capture multivariate correlations and feed-forward networks to learn series representations. PatchTST~\citep{Nie2023PatchTST} segments time series into subseries-level patches as input tokens and employs channel-independence. The channel-independence strategy improves robustness and adaptability by enabling distinct attention paths for each channel, in contrast to channel-mixing methods. Our HaKAN framework also adopts this channel-independent approach to preserve the unique temporal dynamics of each variable of the multivariate time series. Despite the success of Transformer-based methods in time series forecasting, their self-attention mechanism is, however, permutation-equivariant, meaning that it does not naturally preserve the temporal order, potentially compromising the modeling of time-dependent information.

\textbf{MLP- and KAN-based Models.}\quad Various MLP-based models have been adopted for long-term time series forecasting~\citep{Chen2023TSmixer,Challu2023NHiTS,Wang2024TimeSeriesMLP,Zeng2023DLinear} due to their architectural and computational efficiency. For instance, TSMixer~\citep{Chen2023TSmixer} captures temporal patterns and cross-variate information by interleaving time-mixing and feature-mixing MLPs., while DLinear~\citep{Zeng2023DLinear} enhances long-term time series forecasting by decomposing input data into trend and seasonal components. While MLP-based models offer greater structural simplicity and faster computation compared to Transformer-based models, they often struggle to capture global temporal dependencies and typically require longer input sequences to match the performance of more expressive architectures. More recently, TimeKAN~\citep{huang2025timekan} introduces a KAN-based architecture that decomposes multivariate time series into multiple frequency bands using cascaded frequency decomposition and moving averages. Similarly, TsKAN~\citep{chen2025tskan} presents a KAN-based approach that incorporates a multi-scale patching  module to extract temporal and cross-dimensional features across scales. Our proposed HaKAN framework differs from these KAN-based models by using inter-patch KAN layers to capture global dependencies, overcoming MLPs' reliance on long input sequences, and from Transformer-based models by using an efficient Mixer-like structure that leverages KAN layers with Hahn polynomials for flexible function approximation. Its advantages include effective modeling of global and local temporal patterns, mitigation of spectral bias, and computational efficiency.

\section{METHOD}\label{sec:method}

\begin{figure*}[!htb]
\centering
\includegraphics[scale=0.71]{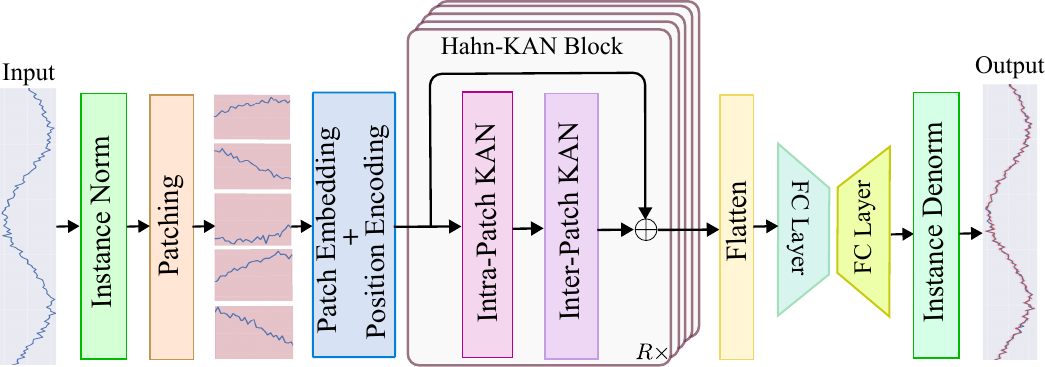}
\caption{\textbf{HaKAN Architecture}. The model integrates channel independence, reversible instance normalization, and patching, followed by patch and position embeddings. A stack of $R$ Hahn-KAN blocks, each with intra-patch and inter-patch KAN layers using Hahn polynomials, processes the embedded sequence to capture temporal patterns. The output is mapped through a bottleneck structure with two fully connected layers to produce the final forecast.}
\label{Fig:KANMixerArch}
\end{figure*}

\subsection{Problem Description and Preliminaries}
\noindent\textbf{Problem Statement.}\quad Time series forecasting refers to the process of predicting future values over a period of time using historical data. Let $\bm{X}_{1:L}=(\bm{x}_{1},\cdots,\bm{x}_{L})^{\T}\in\mathbb{R}^{L\times M}$ be a history sequence of $L$ multivariate time series, where for any time step $t$, each row $\bm{x}_{t}=(x_{t1},\dots,x_{tM})\in\mathbb{R}^{1\times M}$ is a multivariate vector consisting of $M$ variables or channels. The goal of multivariate time series forecasting is to predict a sequence $\hat{\bm{X}}_{L+1:L+T}=(\hat{\bm{x}}_{L+1},\dots,\hat{\bm{x}}_{L+T})^{\T}\in\mathbb{R}^{T\times M}$ for the future $T$ timesteps.

\medskip\noindent\textbf{Kolmogorov-Arnold Networks.}\quad KANs are inspired by the Kolmogorov-Arnold representation theorem~\citep{Braun2009KANTH,Johannes2021KANTH}, which states that any continuous multivariate function on a bounded domain can be represented as a finite composition of continuous univariate functions of the input variables and the binary operation of addition. A KAN layer, a fundamental building block of KANs~\citep{Liu2024KAN}, is defined as a matrix of 1D functions $\bg{\Phi}=(\phi_{q,p})$, where each trainable activation function $\phi_{q,p}$ is defined as a weighted combination, with learnable weights, of a sigmoid linear unit function and a spline function. Given an input vector $\bm{x}$, the output of an $\mathsf{L}$-layer KAN is given by
\begin{equation}
\text{KAN}(\bm{x})=(\bg{\Phi}^{(\mathsf{L}-1)}\circ\dots\circ\bg{\Phi}^{(1)}\circ\bg{\Phi}^{(0)})\bm{x},
\end{equation}
where $\bg{\Phi}^{(\ell)}$ is a matrix of learnable functions associated with the $\ell$-th KAN layer.

\subsection{Proposed HaKAN Framework}
The proposed HaKAN model processes a multivariate time series $\bm{X}_{1:L} \in \mathbb{R}^{L \times M}$ to predict the future sequence $\hat{\bm{X}}_{L+1:L+T} \in \mathbb{R}^{T \times M}$. As illustrated in Figure~\ref{Fig:KANMixerArch}, the model architecture consists of the following key components:

\textbf{Channel Independence.}\quad Channel independence (CI) is a strategy that treats each feature or variable in a multivariate time series separately~\citep{Nie2023PatchTST}. Instead of combining information across channels, this strategy preserves the unique characteristics of each variable by maintaining their independence. Specifically, the input time series $\bm{X}_{1:L}=(\bm{x}_{1},\dots,\bm{x}_{L})^{\T}$ is split into $M$ univariate series $\bm{x}^{(i)}=(x_{1}^{(i)},\dots,x_{L}^{(i)})^{\T}\in\mathbb{R}^{L}$, where $\bm{x}^{(i)}$ is the $i$th column of $\bm{X}_{1:L}$. Each of these univariate series is fed into the model backbone. Our HaKAN model takes $\bm{x}^{(i)}$ as input and returns a $T$-dimensional vector of predictions $\hat{\bm{x}}^{(i)}=(\hat{x}_{L+1}^{(i)},\dots,\hat{x}_{L+T}^{(i)})^{\T}$.

\textbf{Normalization.}\quad Each input series is normalized using the reversible instance normalization (RevIN) technique~\citep{Kim2022RevIN}, which addresses challenges related to shifts in data distributions over time. RevIN consists of two main steps: normalization and denormalization. In the first step, the input undergoes normalization to standardize its distribution in terms of mean and variance. After the model generates output sequences, RevIN reverses the normalization process by denormalizing these outputs.

\textbf{Patching.}\quad Each normalized univariate series is partitioned into a sequence of patches to improve computational efficiency and capture local temporal patterns~\citep{Nie2023PatchTST}. The series is divided into patches $\bm{X}^{(i)}_p \in \mathbb{R}^{N \times P}$, where $P$ is the patch length, $N=\lfloor\frac{L-P}{S}\rfloor+2$ is the number of patches, and $S$ is the stride of the sliding window. Patches are generated by sliding a window of size $P$ over the series with stride $S$. If the last patch has fewer than $P$ time steps, the final time step of the normalized univariate series is repeated to pad the patch. Patching offers several advantages, including improved retention of local semantic information, enhanced computational and memory efficiency, and access to a broader historical context.

\textbf{Patch and Position Embeddings.}\quad Each patch in $\bm{X}^{(i)}_p \in \mathbb{R}^{N \times P}$ is projected into a $D$-dimensional embedding using a temporal linear projection with a trainable weight matrix \( \bm{W}_p \in \mathbb{R}^{P \times D} \). To retain the temporal order of the patches, which is critical for time series forecasting, a learnable positional embedding matrix $\bm{W}_{\text{pos}}\in\mathbb{R}^{N\times D}$ is added:
\begin{equation}
\bm{X}^{(i)}_d = \bm{X}^{(i)}_p \bm{W}_p + \bm{W}_{\text{pos}},
\end{equation}
where $\bm{X}^{(i)}_d \in\mathbb{R}^{N\times D}$ is the embedded sequence for the $i$-th channel. Each row of $\bm{X}^{(i)}_d$, referred to as a temporal patch-level token, represents the embedded features of a single patch from the $i$-th channel, maintaining the channel independence of the CI strategy. The positional embeddings ensure the model captures the sequential nature of the patches, addressing the causal structure of time series data. The embedded sequence serves as input to the Hahn-KAN block.

\textbf{Hahn-KAN Block.}\quad The core component of our model architecture is the Hahn-KAN block, which processes the embedded sequence $\bm{X}^{(i)}_d \in \mathbb{R}^{N\times D}$ to capture both global and local temporal patterns. Each block consists of two KAN layers with Hahn Polynomials, structured with a residual connection:
\begin{equation}
\bm{X}^{(i)}_k = \text{KAN}(\text{KAN}(\bm{X}^{(i)}_d)^\T)^\T + \bm{X}^{(i)}_d,
\end{equation}
where $\bm{X}^{(i)}_k \in\mathbb{R}^{N\times D}$ is the output of the block, and each $\text{KAN}(\cdot)$ operation corresponds to a single KAN layer with univariate functions parameterized by Hahn polynomials. Specifically, each trainable univariate function $\phi_{q,p}$ of the KAN layer is parameterized using Hahn polynomials~\citep{Koekoek2010Poly} to provide flexibility in function approximation:
\begin{equation}
\bm{\phi}_{q, p}(x_{p}) = \sum_{r=0}^{d} \gamma_{q,p,r} P_{r}(x_{p}),
\end{equation}
where $x_p$ represents the $p$-th element of the KAN input vector, and $\gamma_{q,p,r}$ is the learnable coefficient of the $r$-th Hahn polynomial $P_{r}(x_p)$ for the $q$-th output element. The $r$-th Hahn polynomial $P_{r}(x)=\text{Hahn}(a,b,n)$, with parameters $a$, $b$ and $n$, is defined by the recurrence relation
\begin{equation}
A P_r(x) = (A + B - x) P_{r-1}(x) - B P_{r-2}(x),
\end{equation}
with coefficients:
\begin{equation}
A = \frac{(r + a + b)(r + a)(n - r + 1)}{(2r + a + b - 1)(2r + a + b)},
\end{equation}
\begin{equation}
B = \frac{(r - 1)(r + b - 1)(r + a + b + n)}{(2r + a + b - 2)(2r + a + b - 1)},
\end{equation}
and initial conditions $P_0(x) = 1$, $P_1(x) = 1 - \frac{a + b + 2}{(a + 1)n} x$.

The Hahn-KAN block consists of two nested layers: an intra-patch KAN layer (feature-mixing) and an inter-patch KAN layer (patch-mixing), both parameterized by Hahn polynomials. The inter-patch layer focuses on cross-patch relationships to capture global temporal patterns across the entire look-back window, such as patterns spanning the look-back window timesteps, while the intra-patch layer refines the features by focusing on local patterns within each patch. The latter captures fine-grained patterns within each patch, such as sudden changes in a short time window. The residual connection ensures training stability by allowing the Hahn-KAN block to learn incremental updates to the input.

The use of Hahn Polynomials in both intra-KAN and inter-KAN layers enhances the model's ability to approximate complex temporal functions, mitigating the spectral bias of traditional MLPs and providing interpretability through learnable activation functions. To capture hierarchical temporal patterns, the Hahn-KAN block is repeated \( R \) times in a stack, with each block taking the output of the previous block as its input, starting with the embedded sequence \( \bm{X}^{(i)}_d \). The output of the \( r \)-th block, \( \bm{X}^{(i)}_{k,r} \in \mathbb{R}^{N \times D} \), becomes the input to the \((r+1)\)-th block. After $R$ blocks, the final output $\bm{X}^{(i)}_k \in\mathbb{R}^{N\times D}$ is flattened into a feature vector $\bm{x}_{f}^{(i)}\in\mathbb{R}^{ND}$, where $ND$ is the total feature dimension. This stacking mechanism enables the model to iteratively refine the features, capturing patterns at multiple temporal scales, from short-term fluctuations to long-term trends.

\textit{Why KAN with Hahn Polynomials?}\quad In a standard KAN layer with $d_{\text{in}}$-dimensional inputs and $d_{\text{out}}$-dimensional outputs, a B-spline of order $d$ and grid size $G$ is used as a learnable activation function. Unlike standard KANs, our proposed Hahn polynomial-based KANs offer superior computation and parameter efficiency. First, Hahn polynomials eliminate the need for grid discretization, removing the dependency on grid size $G$, a key factor in the complexity of standard KANs. Second, while standard KANs incur a time complexity of $\mathcal{O}(d_{\text{in}}d_{\text{out}}[9d(G + 1.5d) + 2G - 2.5d + 3])$~\citep{Yang2025KAT}, our Hahn KANs achieve a simplified complexity of $\mathcal{O}(d_{\text{in}} d_{\text{out}} d)$, where $d$ is the Hahn polynomial degree (typically $d=3$). This is comparable to the $\mathcal{O}(d_{\text{in}} d_{\text{out}})$ complexity of MLPs. Third, Hahn KANs require only $(d_{\text{in}} d_{\text{out}} (d+1))$ parameters, significantly fewer than the $(d_{\text{in}} d_{\text{out}} (G + d + 3) + d_{\text{out}})$ parameters of standard KANs~\citep{Yang2025KAT}. This efficient design, coupled with polynomial-time evaluation and full parallelizability, makes our proposed HaKAN model a lightweight framework for time series forecasting.

\textbf{Output Layer with Bottleneck Structure.}\quad  The flattened vector $\bm{x}_{f}^{(i)}\in\mathbb{R}^{ND}$ is passed through an output layer consisting of two fully connected layers that form a bottleneck structure, mapping the features to the prediction horizon $T$. The first layer is a down-projection, which reduces the dimensionality of the feature vector to a bottleneck middle dimension $H$, using a weight matrix $\bm{W}_{\text{down}}\in\mathbb{R}^{H\times ND}$:
\begin{equation}
\bm{h}^{(i)} = \bm{W}_{\text{down}}\bm{x}_{f}^{(i)},
\end{equation}
where $\bm{h}^{(i)}\in\mathbb{R}^{H}$. This compression reduces both the risk of overfitting and the computational cost of the output layer.

The second layer is an up-projection, which expands the compressed features to the prediction horizon $T$, using a weight matrix $\bm{W}_{\text{up}}\in\mathbb{R}^{T\times H}$:
\begin{equation}
\hat{\bm{x}}^{(i)} = \bm{W}_{\text{up}}\bm{h}^{(i)},
\end{equation}
where \( \hat{\bm{x}}^{(i)} \in \mathbb{R}^T \) is the forecasted sequence for the \( i \)-th channel.

The bottleneck structure ensures efficient mapping to the prediction horizon, especially for large $T$, by first compressing the features before expanding them. The forecasted sequences for all $M$ channels are combined to form the final output $\hat{\bm{X}}_{L+1:L+T}\in\mathbb{R}^{T\times M}$. Finally, RevIN denormalization is applied to $\hat{\bm{x}}^{(i)}$ for each channel, using the stored mean and standard deviation, to restore the original data scale.

\subsection{Model Training}
The parameters of our HaKAN model are learned by minimizing the following training objective function
\begin{equation}
\mathcal{L}=\frac{1}{MT}\sum_{i=1}^{M}\sum_{\tau=L+1}^{L+T}\big\Vert \bm{x}_{\tau}^{(i)} - \hat{\bm{x}}_{\tau}^{(i)}\big\Vert^2,
\end{equation}
where $\bm{x}_{\tau}^{(i)}$ and $\hat{\bm{x}}_{\tau}^{(i)}$ are the ground-truth and prediction, respectively, $\tau\in\{L+1,\dots,L+T\}$, $L$ is the look-back window, $T$ is the prediction horizon, and $M$ is the number of time series variables.

The main algorithmic steps of the proposed HaKAN framework are summarized in Algorithm 1.
\begin{algorithm}[!h]
	\caption{\small HaKAN: Time series forecasting}
	\small
	\begin{algorithmic}[1]  
		\Require Input multivariate time series $\bm{X}_{1:L} \in \mathbb{R}^{L \times M}$ with look-back $L$ and $M$ channels; forecast horizon $T$
		\Ensure Forecasted sequence $\hat{\bm{X}}_{L+1:L+T} \in \mathbb{R}^{T \times M}$
		
		\For{$i = 1$ to $M$} \Comment{\textit{Channel independence}}
		\State Using RevIN, normalize the channel univariate series $\bm{x}^{(i)} = (\bm{x}^{(i)}_1, \ldots, \bm{x}^{(i)}_L)^\T \in \mathbb{R}^{L}$
		\State Partition the normalized channel univariate series into $N$ patches of size $P$ to generate $\bm{X}^{(i)}_p \in \mathbb{R}^{N \times P}$
		\State Embed patches: $\bm{X}^{(i)}_d = \bm{X}^{(i)}_p \bm{W}_p + \bm{W}_{\text{pos}} \in \mathbb{R}^{N \times D}$
		\State Initialize $\bm{X}^{(i)}_{k} = \bm{X}^{(i)}_d$
		\For{$r = 1$ to $R$} \Comment{\textit{Hahn-KAN blocks}}
		\State $\bm{X}^{(i)}_{k} = \text{KAN}(\text{KAN}(\bm{X}^{(i)}_{k})^\T)^\T + \bm{X}^{(i)}_{k}$
		\EndFor
		\State Flatten $\bm{X}^{(i)}_{k} \in \mathbb{R}^{N \times D} \rightarrow \bm{x}_{f}^{(i)} \in \mathbb{R}^{ND}$
		\State Bottleneck mapping:
		\State \hskip1em $\bm{h}^{(i)} = \bm{W}_{\text{down}} \bm{x}_{f}^{(i)} \in \mathbb{R}^{H}$
		\State \hskip1em $\hat{\bm{x}}^{(i)} = \bm{W}_{\text{up}} \bm{h}^{(i)} \in \mathbb{R}^{T}$
		\State Denormalize $\hat{\bm{x}}^{(i)}$ via RevIN
		\EndFor
		\State Combine the channels: $\hat{\bm{X}}_{L+1:L+T} = (\hat{\bm{x}}^{(1)}, \ldots, \hat{\bm{x}}^{(M)})$
	\end{algorithmic}
\end{algorithm}

\section{EXPERIMENTS}\label{sec:experiments}

\subsection{Experimental Setup}
\textbf{Datasets.}\quad We evaluate HaKAN on several benchmark datasets: Weather, Electricity, Illness, and four ETT datasets (ETTh1, ETTh2, ETTm1, ETTm2)~\citep{Wu2021Autoformer}. \textbf{Weather} records 21 meteorological indicators every 10 minutes throughout 2020. \textbf{Traffic} comprises hourly road occupancy data from sensors across San Francisco Bay area freeways. \textbf{Electricity} tracks hourly electricity usage for 321 customers from 2012 to 2014. \textbf{ETT} includes transformer load and oil temperature data, sampled hourly for ETTh datasets and every 15 minutes for ETTm datasets, spanning July 2016 to July 2018. \textbf{Illness} contains weekly records of patient counts and influenza-like illness ratios.

\textbf{Baselines and Evaluation Metrics.} We evaluate the performance of our model against various recent state-of-the-art methods, including S-Mamba~\citep{Wang2025SMamba},
TimeKAN~\citep{huang2025timekan}, Timer-XL~\citep{liu2025timerxl}, TsKAN~\citep{chen2025tskan}, iTransformer~\citep{Liu2024iTransformer}, PatchTST~\citep{Nie2023PatchTST}, TimesNet~\citep{Wu2023TimesNet}, Crossformer~\citep{Zhang2023Crossformer}, DLinear and RLinear~\citep{Zeng2023DLinear}, N-HiTS~\citep{Challu2023NHiTS}, TiDE~\citep{Das2023TiDe}, MICN~\citep{Wang2023MICN}, and FEDformer~\citep{Zhou2022FEDformer}. PatchTST includes 2 variants, PatchTST/42 and PatchTST/64, with the latter being the best performing model. Performance is evaluated using mean squared error (MSE) and mean absolute error (MAE).

\begin{table*}[!htb]
	\caption{Time series forecasting results across prediction lengths $T\in \{24, 36, 48, 60\}$ for the Illness dataset and $T\in \{96, 192, 336, 720\}$ for the other datasets. The best results are highlighted in \textbf{bold}, and the second-best are \underline{underlined}. For each method, multiple look-backs $L\in\{96, 192, 336, 720\}$ are evaluated, with the best-performing look-back reported. The dash (-) indicates no reported results in the baselines' papers.}
	\smallskip
	\setlength\tabcolsep{2.8pt} 
	\centering
	\scriptsize
	\begin{tabular}{l*{18}{c}}
		\toprule
		Method & \multicolumn{2}{c}{\textbf{HaKAN}} & \multicolumn{2}{c}{TsKAN} & \multicolumn{2}{c}{Timer-XL}
		& \multicolumn{2}{c}{TimeKAN} & \multicolumn{2}{c}{PatchTST/64} & \multicolumn{2}{c}{N-HiTS} & \multicolumn{2}{c}{DLinear} & \multicolumn{2}{c}{MICN} & \multicolumn{2}{c}{TimesNet}\\
	    & \multicolumn{2}{c}{\textbf{(ours)}}
        & \multicolumn{2}{c}{(\citeyear{chen2025tskan})}
        &
        \multicolumn{2}{c}{(\citeyear{liu2025timerxl})} &
         \multicolumn{2}{c}{(\citeyear{huang2025timekan})}
		& \multicolumn{2}{c}{(\citeyear{Nie2023PatchTST})} & \multicolumn{2}{c}{(\citeyear{Challu2023NHiTS})} & \multicolumn{2}{c}{(\citeyear{Zeng2023DLinear})} & \multicolumn{2}{c}{(\citeyear{Wang2023MICN})} & \multicolumn{2}{c}{(\citeyear{Wu2023TimesNet})}\\
		\cmidrule(lr){2-3} \cmidrule(lr){4-5} \cmidrule(lr){6-7} \cmidrule(lr){8-9} \cmidrule(lr){10-11} \cmidrule(lr){12-13} \cmidrule(lr){14-15} \cmidrule(lr){16-17} \cmidrule(lr){18-19}
		Metric & MSE & MAE & MSE & MAE & MSE & MAE & MSE & MAE & MSE & MAE & MSE & MAE & MSE & MAE & MSE & MAE & MSE & MAE\\
		
			\midrule
		\multicolumn{1}{r|}{\parbox[t]{2mm}{\multirow{5}{*}{\rotatebox[origin=c]{90}{ETTh1}}}\quad 96} & 0.369 & \underline{0.394} & 0.376 & 0.395 & \textbf{0.364} & 0.397 & \underline{0.367} & 0.395 & 0.379 & 0.401 & 0.378 & 0.436 & 0.375 & 0.399 & 0.413 & 0.442 & 0.421 & 0.440\\
		\multicolumn{1}{r|}{\quad 192} & \textbf{0.406} & \textbf{0.414} & 0.419 & 0.426 & \underline{0.405} & 0.424 & 0.414 & 0.420 & 0.413 & 0.429 & 0.427 & 0.436 & \underline{0.405} & 0.420 & 0.451 & 0.462 & 0.511 & 0.498\\
		\multicolumn{1}{r|}{\quad 336} & \textbf{0.402} & \textbf{0.421} & 0.449 & 0.450 &  0.427 & 0.439 & 0.445 & 0.434 & 0.435 & 0.436 & 0.458 & 0.484 & \underline{0.439} & \underline{0.443} & 0.556 & 0.528 & 0.484 & 0.478\\
		\multicolumn{1}{r|}{\quad 720} & \textbf{0.443} & \textbf{0.459} & 0.464 & 0.475 &\underline{0.439} & 0.459 & 0.444 & 0.459 & 0.446 & 0.464 & 0.472 & 0.551 & 0.472 & 0.490 & 0.658 & 0.607 & 0.554 & 0.527\\[1ex]
		
		\multicolumn{1}{r|}{\quad Avg.} & \textbf{0.405} & \textbf{0.422} & 0.427 & 0.436 & \underline{0.409} & 0.430 & 0.417 & \underline{0.427} & 0.418 & 0.432 & 0.434 & 0.477 & 0.423 & 0.438 & 0.519 & 0.510 & 0.492 & 0.486 \\
		
		\midrule
		\multicolumn{1}{r|}{\parbox[t]{2mm}{\multirow{5}{*}{\rotatebox[origin=c]{90}{ETTh2}}}\quad 96} & \textbf{0.260} & \textbf{0.328} & 0.282 & 0.342 & \underline{0.277} & 0.343 & 0.290 & 0.340 & 0.274 & 0.337 & 0.274 & 0.345 & 0.289 & 0.353 & 0.303 & 0.364 & 0.366 & 0.417\\
		\multicolumn{1}{r|}{\quad 192} & \textbf{0.319} & \textbf{0.373} & 0.361 & 0.391 & \underline{0.348} & 0.391 & 0.375 & 0.392 & 0.332 & \underline{0.380} & 0.353 & 0.401 & 0.383 & 0.418 & 0.403 & 0.446 & 0.426 & 0.447\\
		\multicolumn{1}{r|}{\quad 336} & \textbf{0.318} & \textbf{0.380} & 0.407 & 0.427 & \underline{0.375} & 0.418 & 0.423 & 0.435 &0.363 & \underline{0.397} & 0.382 & 0.425 & 0.448 & 0.465 & 0.603 & 0.550 & 0.406 & 0.435\\
		\multicolumn{1}{r|}{\quad 720} & 0.394 & 0.432 & 0.415 &0.448 & 0.409 & 0.458 & 0.443 & 0.449 & \textbf{0.393} & \textbf{0.430} & 0.625 & 0.557 & 0.605 & 0.551 & 1.106 & 0.852 & \underline{0.427} & \underline{0.457}\\[1ex]
		\multicolumn{1}{r|}{\quad Avg.} & \textbf{0.323} & \textbf{0.378} & 0.366 & 0.402 & 0.352 & 0.402 & 0.383 & 0.404 & \underline{0.341} & \underline{0.386} & 0.408 & 0.432 & 0.431 & 0.447 & 0.604 & 0.553 & 0.406 & 0.439 \\
		
		\midrule
		\multicolumn{1}{r|}{\parbox[t]{2mm}{\multirow{5}{*}{\rotatebox[origin=c]{90}{ETTm1}}}\quad 96} & \underline{0.289} & \underline{0.345} & 0.310 & 0.356 & \textbf{0.290} & \textbf{0.341} & 0.322 & 0.361 & 0.293 & 0.346 & 0.302 & 0.350 & 0.299 & 0.343 &0.308 &0.360& 0.356 & 0.385\\
		\multicolumn{1}{r|}{\quad 192} & \underline{0.329} & \underline{0.370} & 0.350 & 0.378 & \textbf{0.337} & \textbf{0.369} & 0.357 & 0.383 & 0.333 & 0.370 & 0.347 & 0.383 & 0.335  & 0.365 & 0.343 & 0.384 &0.452 & 0.428\\
		\multicolumn{1}{r|}{\quad 336} & \textbf{0.360} & \textbf{0.391} & 0.368 & 0.394 & 0.374 & 0.392 & 0.382 & 0.401 & \underline{0.369} & \underline{0.392} & \underline{0.369} & 0.402 & \underline{0.369} & 0.386 & 0.395 & 0.411 & 0.419 &0.425\\
		\multicolumn{1}{r|}{\quad 720} & \underline{0.418} & \textbf{0.416} & 0.433 & 0.440 & 0.437 & 0.428 & 0.445 & 0.435 & \textbf{0.416} & \underline{0.420} & 0.431 & 0.441 & 0.425 & 0.421 & 0.427 & 0.434 & 0.452 & 0.451\\[1ex]
		\multicolumn{1}{r|}{\quad Avg.} & \textbf{0.349} & \textbf{0.380} & 0.365 & 0.392 & 0.359 & \underline{0.382} & 0.377 & 0.395 & \underline{0.353} & \underline{0.382} & 0.362 & 0.394 & 0.357 & 0.379 & 0.368 & 0.397 & 0.420 & 0.422 \\

		\midrule
		\multicolumn{1}{r|}{\parbox[t]{2mm}{\multirow{5}{*}{\rotatebox[origin=c]{90}{ETTm2}}}\quad 96} & \textbf{0.166} & \textbf{0.255} & 0.173 & 0.262 & 0.175 & 0.257 & 0.174 & 0.255 & \textbf{0.166} & \underline{0.256} & 0.176 & \textbf{0.255} & \underline{0.167} & 0.260 & 0.169 & 0.268 & 0.188 & 0.276\\
		\multicolumn{1}{r|}{\quad 192} & \textbf{0.222} & \textbf{0.293} & 0.231 & 0.305 & 0.242 & 0.301 & 0.239 & 0.299 & \underline{0.223} & \underline{0.296} & 0.245 & 0.305 & 0.224 & 0.303 & 0.247 & 0.333 & 0.242 & 0.310\\
		\multicolumn{1}{r|}{\quad 336} & \textbf{0.265} & \textbf{0.323} & 0.294 & 0.339 & 0.293 & 0.337& 0.301 & 0.340 & \underline{0.274} & \underline{0.326} & 0.295 & 0.346 & 0.281 & 0.342 & 0.290 & 0.351 & 0.300 & 0.346\\
		\multicolumn{1}{r|}{\quad 720} & \textbf{0.346} & \textbf{0.375} &  0.392 & 0.398 & 0.376 & 0.390 & 0.395 & 0.396 & \underline{0.362} & \underline{0.385} & 0.401 & 0.413 & 0.397 & 0.421 & 0.417 & 0.434 & 0.391 & 0.403\\[1ex]
		\multicolumn{1}{r|}{\quad Avg.} & \textbf{0.250} & \textbf{0.311} & 0.272 & 0.326 & 0.271 & 0.321 & 0.277 & 0.323 & \underline{0.256} & \underline{0.316} & 0.279 & 0.330 & 0.267 & 0.332 & 0.281 & 0.346 & 0.280 & 0.334 \\
		
		\midrule
		\multicolumn{1}{r|}{\parbox[t]{2mm}{\multirow{5}{*}{\rotatebox[origin=c]{90}{Weather}}}\quad 96} & \underline{0.148} & \underline{0.198} & \textbf{0.143} & 0.205 & 0.157 & 0.205 & 0.162 & 0.208 & 0.149 & \underline{0.198} & 0.158 & \textbf{0.195} & 0.176 & 0.237 & 0.178 & 0.249 & 0.163 & 0.219\\
		\multicolumn{1}{r|}{\quad 192} & \textbf{0.190} & \textbf{0.240} & 0.201 & 0.264 & 0.207 & 0.249 & \underline{0.194} & \underline{0.241} & 0.211 & 0.247 & 0.220 & 0.282 & 0.243 & 0.269 & 0.211 & 0.259 & 0.275 & 0.329\\
		\multicolumn{1}{r|}{\quad 336} & \underline{0.242} & \underline{0.282} & 0.256 & 0.301 & 0.259 & 0.291& \textbf{0.206} & \textbf{0.250} & 0.263 & 0.290 & 0.245 & \underline{\textbf{0.282}} & 0.274 & 0.300 & 0.265 & 0.319 & 0.278 & 0.338\\
		\multicolumn{1}{r|}{\quad 720} & 0.317 & \textbf{0.333} & 0.326 & 0.347 & 0.337 & 0.344 & 0.338 & 0.340 & \textbf{0.314} & \underline{0.334} & 0.401 & 0.413 & \underline{0.323} & 0.362 & 0.320 & 0.360 & 0.359 & 0.363\\[1ex]
        \multicolumn{1}{r|}{\quad Avg.} & \textbf{0.224} & \underline{0.263} & 0.231 & 0.279 & 0.240 & 0.272 & \underline{0.225} & \textbf{0.260} & 0.234 & 0.267 & 0.256 & 0.293 & 0.254 & 0.292 & 0.243 & 0.297 & 0.269 & 0.312 \\

		\midrule
		\multicolumn{1}{r|}{\parbox[t]{2mm}{\multirow{5}{*}{\rotatebox[origin=c]{90}{Traffic}}}\quad 96} & 0.365 & 0.252 & -& - & \textbf{0.340} & \textbf{0.238} & -& - & \underline{0.360} & \underline{0.249} & 0.402 & 0.282 & 0.410 & 0.282 & 0.473 & 0.293 & 0.595 & 0.318\\
		\multicolumn{1}{r|}{\quad 192} & 0.391 & 0.262 &  -& - &\textbf{0.360} & \textbf{0.247} & -& - & \underline{0.379} & \underline{0.256} & 0.420 & 0.297 & 0.423 & 0.287 & 0.483 & 0.298 & 0.615 & 0.326\\
		\multicolumn{1}{r|}{\quad 336} & 0.407 & 0.272 &  -& - &\textbf{0.377} & \textbf{0.256} & -& - & \underline{0.392} & \underline{0.264} & 0.448 & 0.313 & 0.436 & 0.296 & 0.491 & 0.303 & 0.616 & 0.326\\
		\multicolumn{1}{r|}{\quad 720} & 0.447 & 0.291 &  -& - & \textbf{0.418} & \textbf{0.279} & -& - &  \underline{0.432} & \underline{0.286} & 0.539 & 0.353 & 0.466 & 0.315 &  0.559 & 0.327 & 0.655 & 0.353\\[1ex]

        \multicolumn{1}{r|}{\quad Avg.} & 0.403 & 0.269 & - & - & \textbf{0.374} & \textbf{0.255} & - & - & \underline{0.391} & \underline{0.264} & 0.452 & 0.311 & 0.434 & 0.295 & 0.502 & 0.305 & 0.620 & 0.331 \\

		\midrule
		\multicolumn{1}{r|}{\parbox[t]{2mm}{\multirow{5}{*}{\rotatebox[origin=c]{90}{Electricity}}}\quad 96}  & \textbf{0.128} & \textbf{0.222} &  -& - &- & - & 0.174 & 0.266 & \underline{0.129} & \textbf{0.222} & 0.147 & 0.249 & 0.140 & \underline{0.237} & 0.157 & 0.266 & 0.178 & 0.284\\
		\multicolumn{1}{r|}{\quad 192} & \textbf{0.146} & \textbf{0.240} &  -& - &- & - &0.182 & 0.273 & \underline{0.147} & \textbf{0.240} & 0.167 & 0.269 & 0.153 & \underline{0.249} & 0.175 & 0.287 & 0.187 & 0.289\\
		\multicolumn{1}{r|}{\quad 336} & \textbf{0.162} & \textbf{0.256} &  -& - &- & - &0.197 & 0.286 & \underline{0.163} & \underline{0.259} & 0.186 & 0.290 & 0.169 & 0.267 & 0.200 & 0.308 & 0.208 & 0.307\\
		\multicolumn{1}{r|}{\quad 720} & \underline{0.202} & \underline{0.292}  & -& - &- & - & 0.236 & 0.320 & \textbf{0.197} & \textbf{0.290} & 0.243 & 0.340 & 0.203 & 0.301 & 0.228 & 0.338 & 0.245 & 0.321\\[1ex]
        \multicolumn{1}{r|}{\quad Avg.} & \underline{0.160} & \textbf{0.253} & - & - & - & - & 0.197 & 0.286 & \textbf{0.159} & \textbf{0.253} & 0.186 & 0.287 & 0.166 & \underline{0.264} & 0.190 & 0.300 & 0.204 & 0.300 \\

		\midrule
		\multicolumn{1}{r|}{\parbox[t]{2mm}{\multirow{5}{*}{\rotatebox[origin=c]{90}{Illness}}}\quad 24} & \textbf{1.183} & \textbf{0.685} &- & - & - & - &- & - & \underline{1.319} & \underline{0.754} & 1.862 & 0.869 & 2.215 & 1.081 & 2.345 & 1.043 & 2.157 & 0.978\\
		\multicolumn{1}{r|}{\quad 36} & \textbf{1.261} & \textbf{0.746} &- & - & - & - &- & - &\underline{1.579} & \underline{0.870} & 2.071 & 0.934 & 1.963 & 0.963 & 2.330 & 1.001 & 2.318 & 1.031\\
		\multicolumn{1}{r|}{\quad 48} & \textbf{1.406} & \underline{0.818} & - & - &- & - &- & - &\underline{1.553} & \textbf{0.815} & 2.134 & 0.932 & 2.130 & 1.024 & 2.386 & 1.051 & 2.121 & 1.005\\
		\multicolumn{1}{r|}{\quad 60} & \underline{1.540} & \underline{0.851} & - & - &- & - &- & - & \textbf{1.470} & \textbf{0.788} & 2.137 & 0.968 & 2.368 & 1.096 & 2.616 & 1.131 & 1.975 & 0.975\\[1ex]
        \multicolumn{1}{r|}{\quad Avg.} & \textbf{1.347} & \textbf{0.775} & - & - & - & - & - & - & \underline{1.480} & \underline{0.807} & 2.051 & 0.926 & 2.169 & 1.041 & 2.419 & 1.056 & 2.143 & 0.997 \\

		\bottomrule
	\end{tabular}
	\label{tab:ComparativeResults}
\end{table*}

\begin{table*}[!h]
	\caption{Long-term time series forecasting results for various prediction lengths $T\in\{96, 192, 336, 720\}$. The look-back is set to 96.}
	\smallskip
	\scriptsize
	\setlength\tabcolsep{3pt}
	\centering
	\begin{tabular}{l*{18}{c}}
		\toprule
		Method & \multicolumn{2}{c}{\textbf{HaKAN}} & \multicolumn{2}{c}{S-Mamba} & \multicolumn{2}{c}{iTransformer} & \multicolumn{2}{c}{RLinear} & \multicolumn{2}{c}{PatchTST/64} & \multicolumn{2}{c}{Crossformer} & \multicolumn{2}{c}{TiDE} & \multicolumn{2}{c}{TimesNet} & \multicolumn{2}{c}{FEDformer} \\
		    & \multicolumn{2}{c}{\textbf{(ours)}}
		& \multicolumn{2}{c}{(\citeyear{Wang2025SMamba})} & \multicolumn{2}{c}{(\citeyear{Liu2024iTransformer})} & \multicolumn{2}{c}{(\citeyear{Zeng2023DLinear})} & \multicolumn{2}{c}{(\citeyear{Nie2023PatchTST})} & \multicolumn{2}{c}{(\citeyear{Zhang2023Crossformer})} & \multicolumn{2}{c}{(\citeyear{Das2023TiDe})} & \multicolumn{2}{c}{(\citeyear{Wu2023TimesNet})} & \multicolumn{2}{c}{(\citeyear{Zhou2022FEDformer})}\\
		\cmidrule(lr){2-3} \cmidrule(lr){4-5} \cmidrule(lr){6-7} \cmidrule(lr){8-9} \cmidrule(lr){10-11} \cmidrule(lr){12-13} \cmidrule(lr){14-15} \cmidrule(lr){16-17} \cmidrule(lr){18-19}
		Metric & MSE & MAE & MSE & MAE & MSE & MAE & MSE & MAE & MSE & MAE & MSE & MAE & MSE & MAE & MSE & MAE & MSE & MAE \\
		\midrule
		\multicolumn{1}{r|}{\parbox[t]{2mm}{\multirow{5}{*}{\rotatebox[origin=c]{90}{ETTh1}}}\quad 96} & \underline{0.383} & \textbf{0.395} & 0.386 & 0.405 & 0.386 & 0.405 & 0.386 & \textbf{0.395} & 0.414 & 0.419 & 0.423 & 0.448 & 0.479 & 0.464 & 0.384 & 0.402  & \textbf{0.376} & 0.419 \\
		\multicolumn{1}{r|}{\quad 192} & \underline{0.434} & \textbf{0.421} & 0.443 & 0.437 & 0.441 & 0.436 & 0.437 & 0.424 & 0.460 & 0.445 & 0.471 & 0.474 & 0.525 & 0.492 & 0.436 & \underline{0.429}  & \textbf{0.420} & 0.448\\
		\multicolumn{1}{r|}{\quad 336} & \underline{0.473} & \textbf{0.439} & 0.489 & 0.468 & 0.487 & 0.458 & 0.479 & \underline{0.446} & 0.501 & 0.466 & 0.570 & 0.546 & 0.565 & 0.515 & 0.491 & 0.469  & \textbf{0.459} & 0.465\\
		\multicolumn{1}{r|}{\quad 720} & \textbf{0.469} & \textbf{0.461} & 0.502 & 0.489 & 0.503 & 0.491 & \underline{0.481} & \underline{0.470} & 0.500 & 0.488 & 0.653 & 0.621 & 0.594 & 0.558 & 0.521 & 0.500 & 0.506 & 0.507\\ [1ex]
		\multicolumn{1}{r|}{\quad Avg.} & \textbf{0.439} & \textbf{0.429} & 0.455 & 0.450 & 0.454 & 0.447 & 0.446 & \underline{0.434} & 0.469 & 0.454 & 0.529 & 0.522 & 0.541 & 0.507 & 0.458 & 0.450 & \underline{0.440} & 0.460 \\
		\midrule
		\multicolumn{1}{r|}{\parbox[t]{2mm}{\multirow{5}{*}{\rotatebox[origin=c]{90}{ETTh2}}}\quad 96}  & \textbf{0.277} & \textbf{0.332} & 0.296 & 0.348 & 0.297 & 0.349 & \underline{0.288} & \underline{0.338} & 0.302 & 0.348 & 0.745 & 0.584 & 0.400 & 0.440 & 0.340 & 0.374 & 0.358 & 0.397 \\
		\multicolumn{1}{r|}{\quad 192 } & \textbf{0.358} & \textbf{0.384} & 0.376 & 0.396 & 0.380 & 0.400 & \underline{0.374} & \underline{0.390} & 0.388 & 0.400 & 0.877 & 0.656 & 0.528 & 0.509 & 0.402 & 0.414 & 0.429 & 0.439\\
		\multicolumn{1}{r|}{\quad 336 } & \textbf{0.342} & \textbf{0.382} & 0.424 & 0.431 & 0.428 & 0.432 & \underline{0.415} & \underline{0.426} & 0.426 & 0.433 & 1.043 & 0.731 & 0.643 & 0.571 & 0.452 & 0.452 & 0.496 & 0.487\\
		\multicolumn{1}{r|}{\quad 720 } & \textbf{0.416} & \textbf{0.436} & 0.426 & 0.444 & 0.427 & 0.445 & \underline{0.420} & \underline{0.440} & 0.431 & 0.446 & 1.104 & 0.763 & 0.874 & 0.679 & 0.462 & 0.468 & 0.463 & 0.474\\ [1ex]
		\multicolumn{1}{r|}{\quad Avg.} & \textbf{0.348} & \textbf{0.383} & 0.381 & 0.405 & 0.383 & 0.407 & \underline{0.374} & \underline{0.398} & 0.387 & 0.407 & 0.942 & 0.684 & 0.611 & 0.550 & 0.414 & 0.427 & 0.437 & 0.449\\
		\multicolumn{1}{r|}{\parbox[t]{2mm}{\multirow{5}{*}{\rotatebox[origin=c]{90}{ETTm1}}}\quad 96} & \textbf{0.328} & \underline{0.368} & 0.333 & \underline{0.368} & 0.334 & \underline{0.368} & 0.355 & 0.376 & \underline{0.329} & \textbf{0.367} & 0.404 & 0.426 & 0.364 & 0.387 & 0.338 & 0.375 & 0.379 & 0.419 \\
		\multicolumn{1}{r|}{\quad 192} & \textbf{0.365} & \textbf{0.385} & 0.376 & 0.390 & 0.377 & 0.391 & 0.391 & 0.392 & \underline{0.367} & \textbf{0.385} & 0.450 & 0.451 & 0.398 & 0.404 & 0.374 & \underline{0.387} & 0.426 & 0.441 \\
		\multicolumn{1}{r|}{\quad 336} & \textbf{0.388} & \textbf{0.404} & 0.408 & 0.413 & 0.426 & 0.420 & 0.424 & 0.415 & \underline{0.399} & \underline{0.410} & 0.532 & 0.515 & 0.428 & 0.425 & 0.410 & 0.411  & 0.445 & 0.459 \\
		\multicolumn{1}{r|}{\quad 720} & \underline{0.457} & \underline{0.442} & 0.475 & 0.448 & 0.491 & 0.459 & 0.487 & 0.450 & \textbf{0.454} & \textbf{0.439} & 0.666 & 0.589 & 0.487 & 0.461 & 0.478 & 0.450 & 0.543 & 0.490 \\ [1ex]
		\multicolumn{1}{r|}{\quad Avg.} & \textbf{0.384} & \textbf{0.399} & 0.398 & 0.405 & 0.407 & 0.410 & 0.414 & 0.407 & \underline{0.387} & \underline{0.400} & 0.513 & 0.496 & 0.419 & 0.419 & 0.400 & 0.406 & 0.448 & 0.452\\
		\midrule
		\multicolumn{1}{r|}{\parbox[t]{2mm}{\multirow{5}{*}{\rotatebox[origin=c]{90}{ETTm2}}}\quad 96} &\underline{0.176} & \underline{0.260} & 0.179 & 0.263 & 0.180 & 0.264 & 0.182 & 0.265 & \textbf{0.175} & \textbf{0.259} & 0.287 & 0.366 & 0.207 & 0.305 & 0.187 & 0.267  & 0.203 & 0.287 \\
		\multicolumn{1}{r|}{\quad 192} & \textbf{0.240} & \textbf{0.301} & 0.250 & 0.309 & 0.250 & 0.309 & 0.246 &0.304 &  \underline{0.241} &  \underline{0.302} & 0.414 & 0.492 & 0.290 & 0.364 & 0.249 & 0.309 & 0.269 & 0.328 \\
		\multicolumn{1}{r|}{\quad 336 } & \textbf{0.299} & \textbf{0.339} & 0.312 & 0.349 & 0.311 & 0.348 & 0.307 & \underline{0.342} & \underline{0.305} & 0.343 & 0.597 & 0.542 & 0.377 & 0.422 & 0.321 & 0.351& 0.325 & 0.366\\
		\multicolumn{1}{r|}{\quad 720} & \textbf{0.392} & \textbf{0.394} & 0.411 & 0.406 & 0.412 & 0.407 & 0.407 & \underline{0.398} & \underline{0.402} & 0.400 & 1.730 & 1.042 & 0.558 & 0.524 & 0.408 & 0.403& 0.421 & 0.415\\ [1ex]
		\multicolumn{1}{r|}{\quad Avg.} & \textbf{0.276} & \textbf{0.324} & 0.288 & 0.332 & 0.288 & 0.332 & 0.286 & 0.327 & \underline{0.281} & \underline{0.326} & 0.757 & 0.610 & 0.358 & 0.404 & 0.291 & 0.333 & 0.305 & 0.349 \\
	\bottomrule
	\end{tabular}
	\label{tab:fixed}
\end{table*}

\textbf{Implementation Details.}\quad All experiments are conducted on a linux machine with a single NVIDIA RTX 4090 GPU 24GB. The HaKAN model is implemented in PyTorch, and Adam~\citep{Kingma2015Adam} is used as optimizer. For the KAN layers, we use Hahn polynomials of the form $\text{Hahn}(a, b, n)$, where $a = 1$, $b = 1$, and $n = 7$, with the polynomial degree fixed at $d = 3$. The number of Hahn-KAN blocks is set to $R = 5$, and the bottleneck dimension is set to $H = 336$. We set a patch length of $P = 16$, a stride of $S = 8$, and a patch embedding dimension to $D = 128$. We follow the standard data partitioning protocols~\citep{Nie2023PatchTST}. Specifically, for the ETT datasets, we use the first 12 months of data for training, the subsequent 4 months for validation, and the final 4 months for testing. This split ensures that if the model fails to generalize to months 13-16, it is unlikely to improve for months 17-20. For the remaining datasets, we adopt a split consisting of 70\% training, 10\% validation, and 20\% testing. HaKAN is trained for up to 100 epochs, with early stopping and patience 10. The learning rate is set to 0.0025 for the Illness dataset, and to 0.0001 for all other datasets.

\subsection{Results and Analysis}
\textbf{Optimized Look-back Window.}\quad To ensure a fair comparison, each baseline is run with look-back windows $L\in\{96, 192, 336, 720\}$, and the best-performing look-back is chosen to avoid underestimating their performance. For the proposed HaKAN model, we similarly evaluate across the same look-back windows and find that the best results are achieved with $L = 336$, which aligns with the optimal look-backs selected for PatchTST~\citep{Nie2023PatchTST} and DLinear~\citep{Zeng2023DLinear}, ensuring consistency in the comparison. All models are evaluated on the Weather, Traffic, Electricity, ETTh1, ETTh2, ETTm1, ETTm2, and Illness datasets for prediction lengths $T\in\{96, 192, 336, 720\}$, using MSE and MAE as evaluation metrics. As reported in Table~\ref{tab:ComparativeResults}, HaKAN consistently outperforms most of the baselines, achieving the best MSE in 18 out of 32 cases and the best MAE in 19 out of 32 cases, with notable relative average MSE and MAE reductions of 8.98\% and 3.96\% on Illness. It excels particularly on datasets with smooth trends like ETT, for instance, achieving a relative average MSE and MAE error reductions of 5.28\% and 2.07\% on ETTh2.

\textbf{Fixed Look-back Window.}\quad A number of baselines, such as S-Mamba~\citep{Wang2025SMamba} and iTransformer~\citep{Liu2024iTransformer}, report the MSE and MAE values for a fixed look-back window of $L = 96$. We also compare our HaKAN model with recent baselines using a fixed look-back $L = 96$. As reported in Table~\ref{tab:fixed}, HaKAN achieves the best average MSE and MAE across prediction lengths \(T \in \{96, 192, 336, 720\}\) on five benchmarks (ETTm1, ETTm2, ETTh1, ETTh2), outperforming strong baselines with notable relative error reductions, though PatchTST and Crossformer remain competitive at shorter horizons. On ETTm1, HaKAN's average MSE/MAE (0.384/0.399) yield relative error reductions of 7.2\%/2.0\% over RLinear, leading at \(T=96, 192, 336\), while PatchTST slightly outperforms at \(T=720\). On ETTm2, HaKAN achieves average MSE/MAE of 0.276/0.324 with relative error reductions of 3.5\%/1.2\%, excelling at \(T=192, 336, 720\), though PatchTST leads at \(T=96\). On ETTh1, HaKAN's average MSE/MAE (0.439/0.429) achieve relative error reductions of 1.6\%/1.2\% over RLinear, leading at \(T=720\) despite FEDformer's advantage at early horizons. On ETTh2, HaKAN dominates with average MSE/MAE (0.348/0.383), offering relative error reductions of 6.9\%/3.8\%, leading across all prediction lengths.

\begin{table*}[!ht]
\centering
\footnotesize 
\begin{minipage}{0.3\textwidth}
\caption{Impact of basis.}
\centering
\scriptsize
\setlength{\tabcolsep}{5pt}
\begin{tabular}{@{}lccc@{}}
\toprule
Basis   & $\overline{\text{MSE}}$ & $\overline{\text{MAE}}$ & Avg.   \\
\midrule
Hahn      & \textbf{0.507} & \textbf{0.431} & \textbf{0.469} \\
Lucas     & 0.531 & 0.435 & 0.482 \\
Chebyshev  & 0.539 & 0.439 & 0.488 \\
B-Splines  & 0.548 & 0.443 & 0.495 \\
\bottomrule
\end{tabular}
\label{tab:poly_basis_performance}
\end{minipage}
\centering
\scriptsize
\setlength{\tabcolsep}{5pt}
\begin{minipage}{0.33\textwidth}
\caption{Impact of number of blocks.}
\centering
\begin{tabular}{lccr}
\toprule
\#Blocks & $\overline{\text{MSE}}$ & $\overline{\text{MAE}}$ & Params (K) \\
\midrule
1    & 0.526 & 0.436 & \textbf{635}\\
3    & 0.534 & 0.438 & 767 \\
5   & \textbf{0.507} & \textbf{0.431} & 899 \\
20  & 0.549 & 0.442 & 1891 \\
\bottomrule
\end{tabular}
\label{tab:num_blocks_performance}
\end{minipage}
\begin{minipage}{0.34\textwidth}
\caption{Impact of hidden dimension.}
\centering
\scriptsize
\setlength{\tabcolsep}{5pt}
\begin{tabular}{lccr}
\toprule
$H$ & $\overline{\text{MSE}}$ & $\overline{\text{MAE}}$ & Params (K) \\
\midrule
200        & 0.536 & 0.438 & 695 \\
336        & \textbf{0.507} & \textbf{0.431} & 899 \\
800        & 0.523 & 0.435 & 1598 \\
1000       & 0.543 & 0.440 & 1899 \\
\bottomrule
\end{tabular}
\label{tab:mid_dim_performance}
\end{minipage}
\end{table*}

\subsection{Ablation Study}\label{Sec:ablation}
In the ablation experiments, we consider six datasets $\mathcal{D}$ = \{ETTh1, ETTh2, ETTm1, ETTm2, Weather, Illness\} and four prediction horizons $\mathcal{T}=\{96, 192, 336, 720\}$ for the first five datasets and $\mathcal{T}=\{24, 36, 48, 60\}$ for Illness. Given a look-back window $L$, we define the average MSE and MAE over all datasets and across all prediction horizons as follows:
\begin{equation}
\footnotesize \overline{\text{MSE}} = \frac{1}{\vert\mathcal{T}\vert \vert\mathcal{D}\vert}\sum_{\delta\in\mathcal{D}}\sum_{T\in\mathcal{T}} \text{MSE}(\bm{X}_{L+1:L+T}^{\delta},\hat{\bm{X}}_{L+1:L+T}^{\delta}),
\end{equation}
\begin{equation}
\footnotesize \overline{\text{MAE}} = \frac{1}{\vert\mathcal{T}\vert \vert\mathcal{D}\vert}\sum_{\delta\in\mathcal{D}}\sum_{T\in\mathcal{T}} \text{MSE}(\bm{X}_{L+1:L+T}^{\delta},\hat{\bm{X}}_{L+1:L+T}^{\delta}),
\end{equation}
where $\bm{X}_{L+1:L+T}^{\delta}$ and $\hat{\bm{X}}_{L+1:L+T}^{\delta}$ are the ground-truth and predicted sequences, respectively, for the dataset $\delta\in\mathcal{D}$.

\textbf{Polynomial Basis.}\quad We conduct an ablation study to assess how different polynomial bases~\citep{Koekoek2010Poly} affect the performance of HaKAN, with results summarized in Table~\ref{tab:poly_basis_performance}. The findings show that the choice of basis functions significantly impacts forecasting performance, with Hahn outperforming alternatives across all metrics.

\textbf{Number of Hahn-KAN Blocks.}\quad Table~\ref{tab:num_blocks_performance} demonstrates a clear trade-off between model performance and parameter efficiency (measured in thousands of learnable parameters) across $R\in\{1,3,5,20\}$. The configuration with $R=5$ provides the best balance, achieving the lowest errors.

\textbf{Bottleneck Dimension.}\quad The bottleneck dimension controls the number of parameters introduced by the down- and up-projection layers in the bottleneck structure of HaKAN. Table~\ref{tab:mid_dim_performance} summarizes the results, which indicate that a bottleneck dimension of $336$ provides the best balance between model size and predictive performance.

\begin{figure}[!htb]
\centering
\includegraphics[width=3in, height=2in]{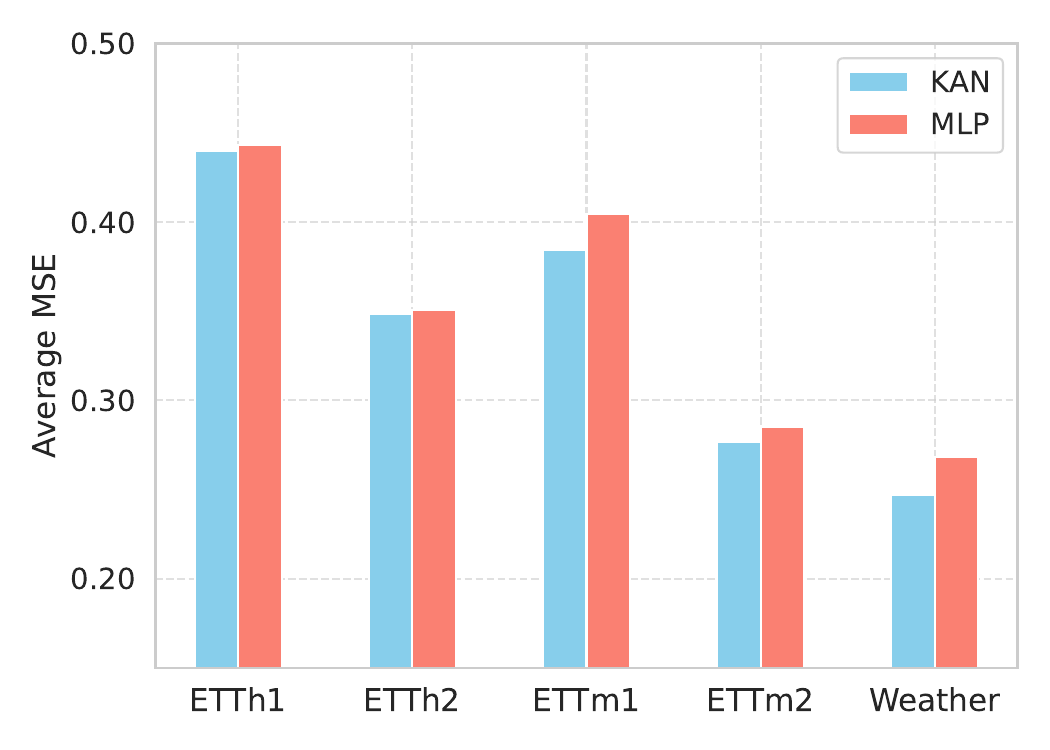}
\caption{Performance comparison between HaKAN and its MLP-based variant across multiple datasets. The look-back window is fixed at $L = 96$, and the average MSE over prediction horizons $T\in\{96, 192, 336, 720\}$ is used as the evaluation metric.}
\label{Fig:MixerBlockEffect}
\end{figure}

\textbf{HaKAN vs. MLP-Based Variant.}\quad Figure~\ref{Fig:MixerBlockEffect} provides a comparative analysis of the forecasting performance of HaKAN against its MLP-based counterpart, where each KAN layer in the HaKAN block is replaced with a fully connected layer. The comparison is conducted across five datasets, with the look-back window fixed at $L=96$, and the average MSE over prediction horizons $T\in\{96, 192, 336, 720\}$ is used as the evaluation metric. The figure shows that HaKAN consistently outperforms the MLP-based variant across all datasets, achieving the lowest average MSE. KAN layers parameterized with Hahn polynomials offer expressive, learnable activation functions for modeling complex temporal dynamics, making our model an efficient alternative to MLP-based models for long-term time series forecasting.

\textbf{Intra-Patch and Inter-Patch.}\quad Table~\ref{Tab:CoreComponents} evaluates the individual and combined contributions of the intra-patch and inter-patch KAN layers, which form the cornerstone of our network architecture, using six ablation datasets with a fixed look-back window of $L=96$ timesteps. This analysis highlights how each component influences the model's ability to capture local and global temporal patterns, respectively. The complete model, integrating both intra-patch and inter-patch layers, achieves the best overall performance, with lowest average MSE and MAE errors, demonstrating the synergistic effect of these components. Notably, removing the intra-patch KAN layer results in the most substantial performance degradation, with errors rising to 0.559 (MSE) and 0.447 (MAE), highlighting its critical role in refining local feature representations within patches. Conversely, omitting the inter-patch KAN layer increases errors to 0.520 (MSE) and 0.435 (MAE), indicating its importance in modeling cross-patch relationships for global context, though the impact is less severe than the intra-patch removal. These findings emphasize the hierarchical design's effectiveness, where the intra-patch layer's focus on fine-grained patterns complements the inter-patch layer's broader temporal perspective, enhancing the model's forecasting accuracy across diverse datasets.

\begin{table}[!htb]
\caption{Effect of intra- and inter-patch KAN layers on model performance.}
\label{Tab:CoreComponents}
\smallskip
\setlength\tabcolsep{9pt} 
\centering
\begin{tabular}{cccc}
\toprule
\multicolumn{2}{c}{Component} & \multicolumn{2}{c}{Metric} \\
\cmidrule(lr){1-2} \cmidrule(lr){3-4}
Intra-Patch & Inter-Patch  & $\overline{\text{MSE}}$ & $\overline{\text{MAE}}$\\
\midrule
\xmark & \cmark    & 0.559 & 0.447 \\
\cmark & \xmark    & 0.520 & 0.435 \\
\cmark & \cmark   & \textbf{0.507} & \textbf{0.431} \\
\bottomrule
\end{tabular}
\end{table}

\begin{figure}[!htb]
\centering
\includegraphics[width=3.1in, height=2in]{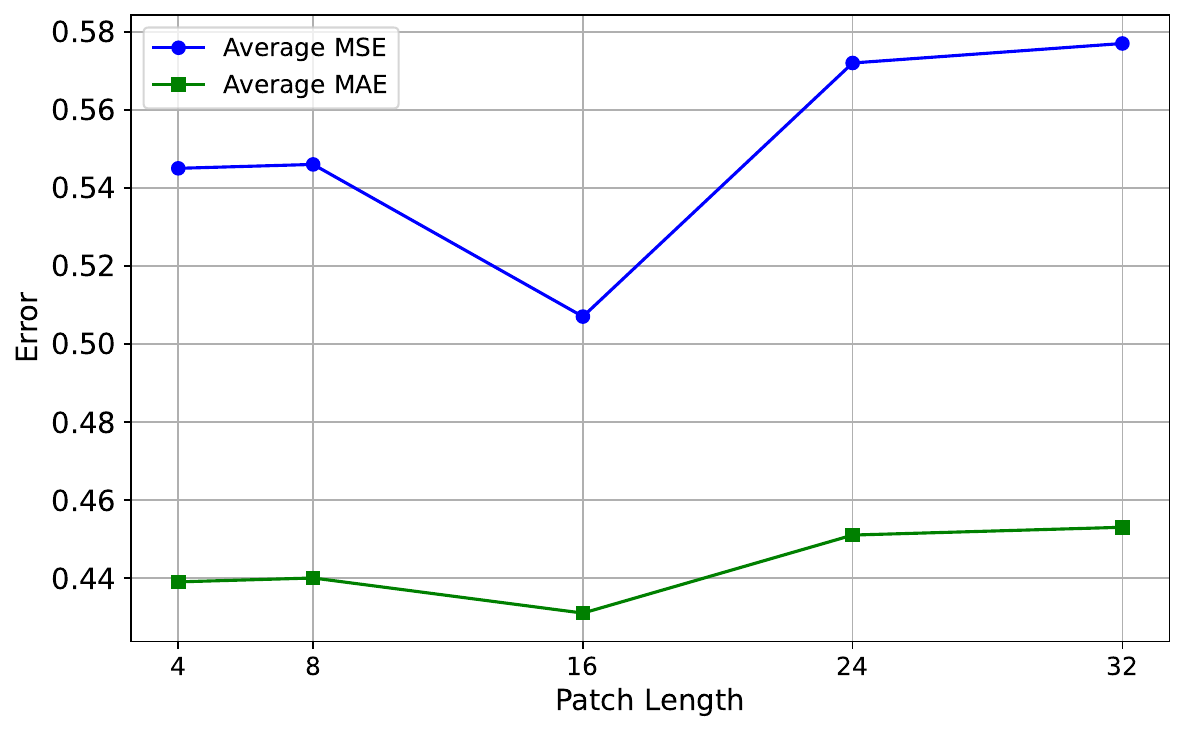}
\caption{Average MSE and MAE results across the six ablation datasets for a varying patch length.}
\label{Fig:PatchLength}
\end{figure}

\textbf{Patch Length.}\quad Figure~\ref{Fig:PatchLength} displays the average MSE and MAE errors for varying patch length $P \in \{4, 8, 16, 24, 32\}$ across all the six ablation datasets. In this experiment, the look-back window is fixed at $L=96$ timesteps, and the stride $S$ is dynamically set to $S = P/2$ to ensure overlapping patches that balance computational efficiency and temporal coverage. The results, depicted in the figure, show that our model achieves the best performance with a patch length of $P=16$, where both average MSE and MAE reach their lowest values, indicating the best trade-off between local pattern capture and global context preservation. This optimal setting suggests that $P=16$ effectively segments the time series into patches that are sufficiently detailed to capture fine-grained temporal dynamics while maintaining enough overlap (stride $S=8$) to support robust forecasting across the datasets.

\subsection{Computational Complexity Analysis}
\textit{Time Complexity.}\quad For each channel, the temporal embedding requires $\mathcal{O}(NPD)$ operations. Each Hahn-KAN block comprises an intra-patch KAN layer with time complexity $\mathcal{O}(ND^2)$ and an inter-patch KAN layer with time complexity $\mathcal{O}(N^2D)$, yielding a total of $\mathcal{O}(R(N^2D + ND^2))$ for $R$ blocks per channel. The bottleneck head adds $\mathcal{O}(NDH + HT)$ operations per channel. As channels are processed independently, the overall time complexity is $\mathcal{O}(M[R(N^2D + ND^2) + NDH + HT])$. In contrast, a Transformer-based encoder incurs a time complexity of $\mathcal{O}(M[RL^2D + NDH + HT])$, dominated by $\mathcal{O}(L^2D)$ for the self-attention term, making HaKAN more efficient, especially for long sequences where $N \ll L$, due to patching and the compact Hahn polynomial representation.

\textit{Space Complexity.}\quad HaKAN stores parameters for each intra-patch KAN layer ($\mathcal{O}(D^{2}(d+1))$ per block) and inter-patch KAN layer ($\mathcal{O}(N^2(d+1))$ per block), totaling $\mathcal{O}(R(N^2(d+1) + D^{2}(d+1)))$ per channel, plus $\mathcal{O}(NDH + HT)$ for the bottleneck head, $\mathcal{O}(PD)$ for patch embedding, $\mathcal{O}(ND)$ for positional encoding, $\mathcal{O}(M)$ for RevIN, and $\mathcal{O}(MND)$ for activation memory. Thus, the simplified total space complexity is $\mathcal{O}(M[R(N^2 + D^{2}) + NDH + HT])$.

\section{DISCUSSION}\label{sec:discussion}
While our proposed HaKAN framework builds upon established concepts, its novelty lies in the integration of Hahn polynomial-based KAN layers within a hierarchical patch-based architecture. The primary motivation for Hahn polynomials is computational efficiency without sacrificing expressiveness. Unlike B-splines, which require grid discretization and incur complexity dependent on grid size, Hahn polynomials eliminate grid dependency, offer closed-form recurrence relations for fast evaluation. HaKAN introduces a \textit{unique integration and optimization strategy} that differentiates it from prior works in three key aspects: (1) \textit{Hahn Polynomial Parameterization:} Unlike TimeKAN~\citep{huang2025timekan} and TsKAN~\citep{chen2025tskan}, which rely on frequency decomposition or spline-based activations, HaKAN leverages Hahn polynomials, an orthogonal basis on discrete domains, providing global approximation capability and parameter efficiency without grid discretization; (2) \textit{Dual-Layer Hahn-KAN Block:} HaKAN introduces a novel hierarchical design with intra-patch and inter-patch KAN layers, enabling simultaneous modeling of fine-grained local patterns and global temporal dependencies; and (3) \textit{Lightweight Complexity Profile:} HaKAN achieves near-MLP efficiency while retaining KAN's flexibility and and interpretability.

Despite its effectiveness, HaKAN assumes channel independence, which limits its performance on datasets with strong inter-variable correlations such as Traffic.

\section{CONCLUSION}\label{sec:conclusion}
In this paper, we introduced HaKAN, a novel framework for multivariate time series forecasting that leverages Kolmogorov-Arnold Networks with Hahn polynomials, effectively capturing both local and global temporal patterns while maintaining computational efficiency. Comparative experiments on several benchmark datasets demonstrated that HaKAN consistently outperforms state-of-the-art baselines across various prediction horizons. This superior performance can be attributed to the KAN layers, which enable the model to approximate complex temporal functions more effectively than MLP- or Transformer-based architectures. Our ablation studies also confirmed the efficacy of key design choices, which collectively minimize forecasting error while balancing model complexity. For future work, we plan to explore the integration of HaKAN with frequency-domain techniques to further enhance its ability to model periodic patterns.

\medskip\noindent\textbf{Acknowledgments.} This work was supported in part by NSERC Discovery and FRQNT Team Grants.

\bibliography{References}

\section*{Checklist}

%

\begin{enumerate}
	
	\item For all models and algorithms presented, check if you include:
	\begin{enumerate}
		\item A clear description of the mathematical setting, assumptions, algorithm, and/or model. [Yes]
		\item An analysis of the properties and complexity (time, space, sample size) of any algorithm. [Yes]
		\item (Optional) Anonymized source code, with specification of all dependencies, including external libraries. [Yes]
	\end{enumerate}
	
	\item For any theoretical claim, check if you include:
	\begin{enumerate}
		\item Statements of the full set of assumptions of all theoretical results. [Not Applicable]
		\item Complete proofs of all theoretical results. [Not Applicable]
		\item Clear explanations of any assumptions. [Not Applicable]
	\end{enumerate}
	
	\item For all figures and tables that present empirical results, check if you include:
	\begin{enumerate}
		\item The code, data, and instructions needed to reproduce the main experimental results (either in the supplemental material or as a URL). [Yes]
		\item All the training details (e.g., data splits, hyperparameters, how they were chosen). [Yes]
		\item A clear definition of the specific measure or statistics and error bars (e.g., with respect to the random seed after running experiments multiple times). [Yes]
		\item A description of the computing infrastructure used. (e.g., type of GPUs, internal cluster, or cloud provider). [Yes]
	\end{enumerate}
	
	\item If you are using existing assets (e.g., code, data, models) or curating/releasing new assets, check if you include:
	\begin{enumerate}
		\item Citations of the creator If your work uses existing assets. [Not Applicable]
		\item The license information of the assets, if applicable. [Not Applicable]
		\item New assets either in the supplemental material or as a URL, if applicable. [Not Applicable]
		\item Information about consent from data providers/curators. [Not Applicable]
		\item Discussion of sensible content if applicable, e.g., personally identifiable information or offensive content. [Not Applicable]
	\end{enumerate}
	
	\item If you used crowdsourcing or conducted research with human subjects, check if you include:
	\begin{enumerate}
		\item The full text of instructions given to participants and screenshots. [Not Applicable]
		\item Descriptions of potential participant risks, with links to Institutional Review Board (IRB) approvals if applicable. [Not Applicable]
		\item The estimated hourly wage paid to participants and the total amount spent on participant compensation. [Not Applicable]
	\end{enumerate}
	
\end{enumerate}

\clearpage
\appendix
\thispagestyle{empty}
\renewcommand{\thesection}{\Alph{section}}
\renewcommand{\thesubsection}{\Alph{section}.\arabic{subsection}}
\onecolumn
\setcounter{page}{1}
\aistatstitle{Supplementary Material}

\section{DATASET DETAILS}
Dataset statistics are summarized in Table~\ref{Tab:datasets}.
\begin{table}[!htb]
\caption{Summary statistics of benchmark datasets.}
\smallskip
\centering
\begin{tabular}{lrrr}
\toprule
Dataset & Features & Timesteps & Frequency \\
\midrule
Weather & 21 & 52,696 & 10 min \\
Traffic & 862 & 17,544 & 1 hour \\
Electricity & 321 & 26,304 & 1 hour \\
Illness & 7 & 966 & 1 week\\
ETTh1, ETTh2 & 7 & 17,420 & 1 hour \\
ETTm1, ETTm2 & 7 & 69,680 & 15 min \\
\bottomrule
\end{tabular}
\label{Tab:datasets}
\end{table}

\section{MODEL ROBUSTNESS AGAINST RANDOM SEEDS}
To evaluate the robustness of our model across different random seeds, we conducted experiments using the seeds $\{2021, 2022, 2023\}$ and calculated the average and standard deviation of MSE and MAE across three runs.

\begin{table}[!htb]
\centering
\caption{Average $\pm$ std of forecasting results (3 seeds) per dataset and horizon.}
\label{tab:avg-std-blocks}
\begin{tabular}{lcc|lcc}
\toprule
\textbf{ETTh1} & MSE & MAE & \textbf{ETTh2} & MSE & MAE \\
\midrule
96  & $0.3663 \pm 0.0015$ & $0.3917 \pm 0.0012$ & 96  & $0.2610 \pm 0.0000$ & $0.3277 \pm 0.0006$ \\
192 & $0.4047 \pm 0.0012$ & $0.4097 \pm 0.0049$ & 192 & $0.3163 \pm 0.0006$ & $0.3673 \pm 0.0006$ \\
336 & $0.4210 \pm 0.0010$ & $0.4243 \pm 0.0012$ & 336 & $0.3077 \pm 0.0006$ & $0.3697 \pm 0.0006$ \\
720 & $0.4490 \pm 0.0026$ & $0.4620 \pm 0.0026$ & 720 & $0.3887 \pm 0.0025$ & $0.4277 \pm 0.0015$ \\
\midrule
\textbf{ETTm1} &  &  & \textbf{ETTm2} &  &  \\
\midrule
96  & $0.2903 \pm 0.0032$ & $0.3460 \pm 0.0017$ & 96  & $0.1670 \pm 0.0000$ & $0.2557 \pm 0.0012$ \\
192 & $0.3287 \pm 0.0015$ & $0.3700 \pm 0.0010$ & 192 & $0.2230 \pm 0.0017$ & $0.2943 \pm 0.0023$ \\
336 & $0.3587 \pm 0.0012$ & $0.3897 \pm 0.0015$ & 336 & $0.2773 \pm 0.0015$ & $0.3293 \pm 0.0015$ \\
720 & $0.4207 \pm 0.0031$ & $0.4210 \pm 0.0056$ & 720 & $0.3757 \pm 0.0090$ & $0.3870 \pm 0.0000$ \\
\midrule
\textbf{Weather} &  &  & \textbf{Electricity} &  &  \\
\midrule
96  & $0.1477 \pm 0.0006$ & $0.1977 \pm 0.0006$ & 96  & $0.1280 \pm 0.0000$ & $0.2227 \pm 0.0006$ \\
192 & $0.1897 \pm 0.0006$ & $0.2400 \pm 0.0000$ & 192 & $0.1460 \pm 0.0000$ & $0.2393 \pm 0.0006$ \\
336 & $0.2420 \pm 0.0000$ & $0.2807 \pm 0.0012$ & 336 & $0.1620 \pm 0.0000$ & $0.2560 \pm 0.0000$ \\
720 & $0.3173 \pm 0.0015$ & $0.3330 \pm 0.0000$ & 720 & $0.2027 \pm 0.0006$ & $0.2920 \pm 0.0000$ \\
\bottomrule
\end{tabular}

\end{table}

\section{HYPERPARAMETERS}
For both look-backs $L=336$ and $L=96$, the number of Hahn-KAN blocks and maximum degree of Hahn polynomials are set to 3. For Hahn polynomial basis $\text{Hahn}(a,b,n)$, we set $a = 1, b = 1, n = 7$.
\begin{table}[!htb]
\centering
\caption{Hyperparameter configurations for each dataset. All experiments used a fixed look-back $L=336$ (except for Illness: $L=104$).}
\label{tab:hparams}
\begin{tabular}{lccccccc}
\toprule
Dataset & $D$ & Patch Length & Stride & Batch Size & Learning Rate & Training Epochs \\
\midrule
\textbf{Electricity} & 128 & 16 & 8 & 32   & 1e-4   & 100 \\
\textbf{ETTh1}       & 128  & 16 & 8 & 256  & 1e-4   & 100 \\
\textbf{ETTh2}       & 128  & 16 & 8 & 256  & 1e-4   & 50  \\
\textbf{ETTm1}       & 128 & 16 & 8 & 1024 & 1e-4   & 100 \\
\textbf{ETTm2}       & 128 & 16 & 8 & 1024 & 1e-4   & 100 \\
\textbf{Illness}     & 16  & 24 & 2 & 64   & 2.5e-3 & 100 \\
\textbf{Traffic}     & 128 & 16 & 8 & 6    & 1e-4   & 100 \\
\textbf{Weather}     & 128 & 16 & 8 & 256  & 1e-4   & 100 \\
\bottomrule
\end{tabular}
\end{table}

\begin{table}[!htb]
\centering
\caption{HaKAN hyperparameter configurations per dataset. Default look-back is $L=96$.}
\label{tab:hparams-hpkan}
\begin{tabular}{lccccccc}
\toprule
Dataset & $D$ & Patch Length & Stride & Batch Size & Learning Rate & Training Epochs \\
\midrule
\textbf{ETTh1}       & 128  & 16 & 8 & 128 & 1e-4  & 100 \\
\textbf{ETTh2}       & 128 & 16 & 8 & 512 & 1e-4  & 50  \\
\textbf{ETTm1}       & 128 & 16 & 8 & 700 & 1e-4  & 100 \\
\textbf{ETTm2}       & 128  & 16 & 8 & 128 & 1e-4  & 100 \\
\bottomrule
\end{tabular}
\end{table}

\section{ADDITIONAL ABLATION STUDIES}
\begin{figure}[!htb]
\centering
\includegraphics[width=.73\linewidth]{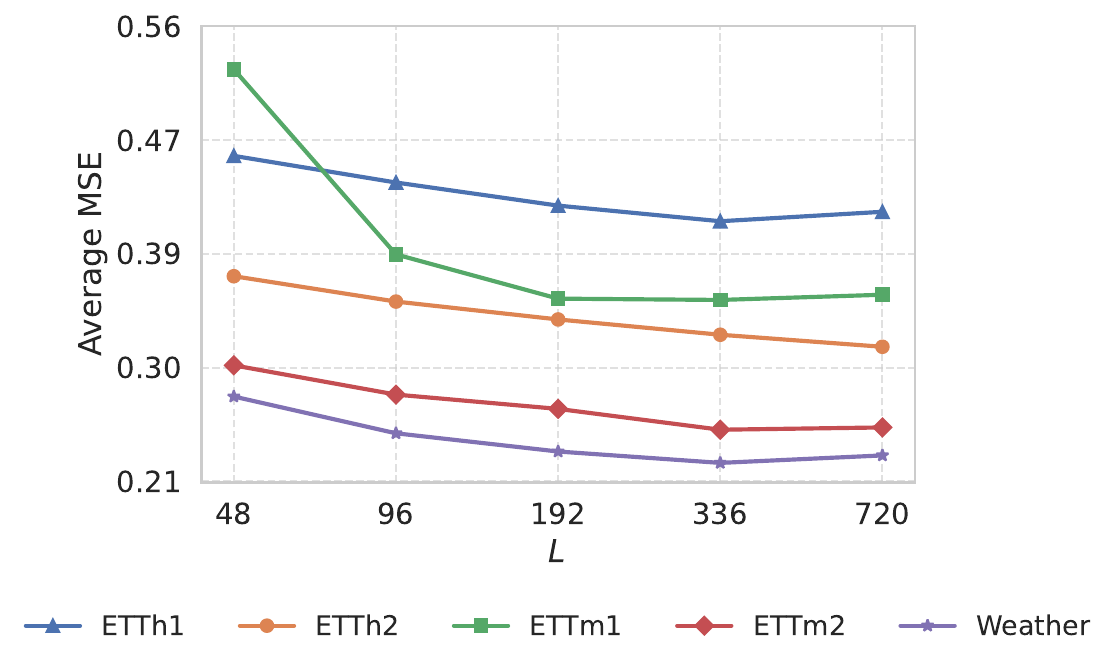}
\caption{Evaluation of long-term forecasting performance across different look-back window lengths on multiple datasets, using the average MSE over prediction horizons $T\in\{96, 192, 336, 720\}$ as the evaluation metric.}
\label{Fig:LookBackEffect}
\end{figure}

\textbf{Effect of Look-back Window.}\quad The look-back window $L$ plays an important role in the HP-KAN model's ability to capture temporal dependencies for long-term time series forecasting, having a direct impact on the number of model parameters due to the use of fixed patch size $P$ and stride $S$. As $L$ increases, the number of patches $N=\lfloor\frac{L-P}{S}\rfloor+2$ also grows, resulting in a larger input sequence. Figure~\ref{Fig:LookBackEffect} illustrates the effect of varying look-back window lengths on the long-term forecasting performance of HP-KAN, with the average MSE across prediction horizons $T \in \{96, 192, 336, 720\}$ for each dataset. The figure reveals that performance consistently improves as $L$ increases from 48 to 336, with the lowest average MSE achieved at $L=336$, reflecting the benefit of increased temporal context.

\vfill

\end{document}